\documentclass[runningheads]{llncs}

\usepackage{eccv}

\usepackage{eccvabbrv}

\usepackage{graphicx}
\usepackage{booktabs}
\usepackage{multirow}
\usepackage{wrapfig}
\usepackage{pifont}
\usepackage{marvosym}

\newcommand{\cmark}{\ding{51}}

\usepackage[accsupp]{axessibility}  

\usepackage{hyperref}

\usepackage{orcidlink}

\begin{document}

\title{RASA: Disentangled Spatial-Motional Priors for Cross-Identity Character Animation}
\titlerunning{Reference-Aware Structural Alignment}

\author{Zhen Xiao\inst{1}\thanks{\scriptsize Work done during internship at HiDream.ai.} \orcidlink{0009-0003-9885-5217} \and
Zhen Shen\inst{2}\orcidlink{0009-0009-8576-3309} \and 
Zhaofan Qiu\inst{2}\orcidlink{0000-0002-7485-9198} \and
Ting Yao\inst{2}\orcidlink{0000-0001-7587-101X} \and  \\
Xueliang Liu\inst{1}\orcidlink{0000-0003-0077-9715} \and 
Tao Mei\inst{2}\orcidlink{0000-0002-5990-7307}
}

\authorrunning{Z.~Xiao et al.}

\institute{Hefei University of Technology, Hefei, China \and HiDream.ai Inc., Hefei, China\\
\email{2023030055@mail.hfut.edu.cn}, \email{\{shenzhen, qiuzhaofan, tiyao\}@hidream.ai}\\
\email{liuxueliang@hfut.edu.cn}, \email{tmei@hidream.ai}}

\maketitle

\begin{abstract}

Cross-identity character animation aims to drive a target identity (from a reference image) to follow the motion of a source character (from a driving video). The core challenge lies in the inherent entanglement of two essential capabilities: cross-identity spatial mapping-aligning position, scale, and skeletal proportions between the reference and the driving pose-and subsequent motion control-refining joint articulation, volumetric consistency, and view coherence during generation. In this paper, we introduce \textbf{Reference-Aware Structural Alignment (RASA)}, a novel framework that systematically disentangles spatial mapping from motion control by injecting structured priors into a Diffusion Transformer (DiT). Our approach operates in two complementary stages. First, a \textbf{Spatial Prior Calibrator (SPC)} fuses the reference identity with the driving pose to generate an initial noise latent that is spatially grounded—it ensures the target character is correctly positioned, scaled, and proportionally aligned with the driving skeleton. This resolves cross-identity spatial mismatches at the very start of generation. Second, to achieve identity-agnostic motion control, we propose an \textbf{Inherent Motional Guider (IMG)}. Moving beyond appearance-biased 2D keypoints, IMG encodes shape-agnostic SMPL articulation parameters into a semantic motion vector. Injected into the intermediate layers of the DiT, this vector serves as complementary guidance that works in tandem with the base pose condition, providing anatomically consistent joint articulation and view-aware volumetric refinement. To rigorously evaluate this challenging task, we curate \textbf{CIM-Bench}, a high-quality benchmark with rigorous manual curation. Extensive experiments demonstrate that RASA significantly outperforms state-of-the-art methods in both motion fidelity and visual quality. Our work establishes a new paradigm for cross-identity animation, showing that disentangled spatial and motional priors are key to achieving robust and consistent character animation. Our project page is at \url{https://hidream-ai.github.io/RASA/}.

\end{abstract}

\begin{figure}[t]
  \centering
  \includegraphics[width=\linewidth]{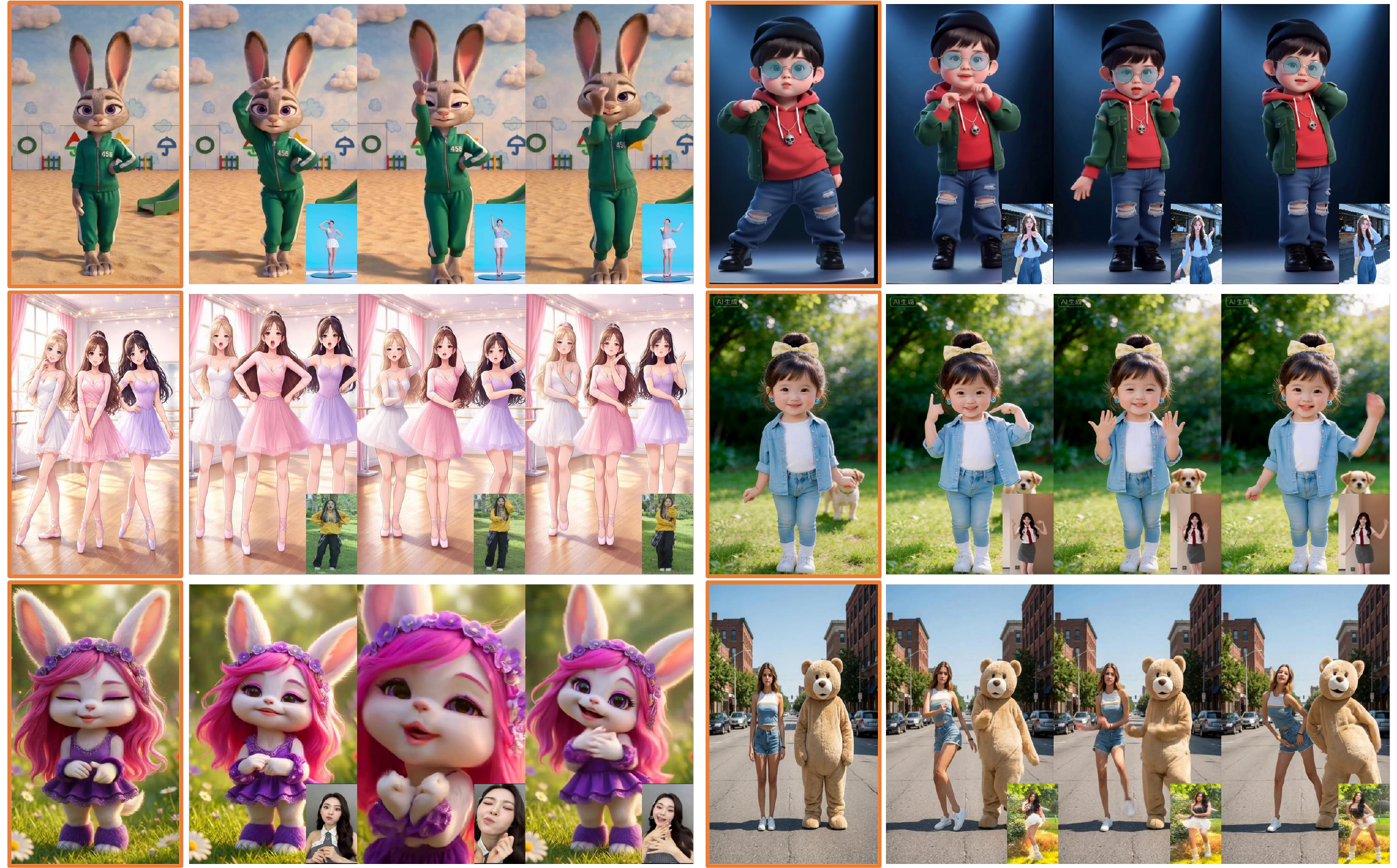}
  \caption{RASA enables high-fidelity character animation under diverse and structurally misaligned driving poses, including large motion variations, stylized identities, and multi-character animation.}
\label{img:fig_1}
\end{figure}

\section{Introduction}
\label{sec:intro}

Character image animation has undergone a paradigm shift with the emergence of Latent Diffusion Models (LDMs)~\cite{rombach2022high} and Diffusion Transformers (DiTs)~\cite{peebles2023scalable}. This capability has unlocked vast potential in digital human synthesis~\cite{hu2024animate,cheng2025wan}, virtual production~\cite{tan2025animate,jiang2025vace}, and interactive media~\cite{hu2025animate,gan2025humandit}, where maintaining structural integrity while following complex motion dynamics is paramount. 

Despite this progress, achieving high-fidelity motion transmission in \textbf{cross-identity scenarios} remains an open challenge. In practice, the driving subject and the reference character often exhibit significant discrepancies in scale, global position, and skeletal proportions. We refer to this as \textbf{pose--reference misalignment}. Current diffusion pipelines typically operate under the implicit assumption of geometric congruence, causing them to struggle when faced with structural mismatches. Traditional remedies—such as explicit pose retargeting~\cite{wang2025unianimate, zhu2024champ, tu2025stableanimator}, pose augmentation~\cite{tan2024animate}, or late-stage feature fusion~\cite{zhang2025steadydancer}—often fall short. Preprocessing-based methods risk information loss, while feature fusion in deep denoising layers inadvertently entangles structural calibration with appearance rendering, leading to visual artifacts and unstable motion transfer.

To overcome these limitations, we propose \textbf{RASA}, a unified framework that resolves pose--reference misalignment through a hierarchical synergy of disentangled spatial and motional priors. Our core insight is that \textbf{spatial mapping} and \textbf{fine-grained motion execution} should be systematically decoupled: geometric discrepancies must be neutralized at the generative onset, while complex motion semantics require deep, consistent guidance. By integrating these two perspectives, RASA enables robust character animation even under extreme identity variations, effectively bridging the gap between 2D spatial layout and 3D anatomical movement, as illustrated in Fig.~\ref{img:fig_1}.

Technically, RASA employs a dual-injection strategy within the DiT backbone. First, we introduce the \textbf{Spatial Prior Calibrator (SPC)} module. By performing feature-level fusion between reference and driving poses, SPC generates a structural prior that is integrated into the latent representations at each denoising step, continuously rectifying 2D spatial mismatches throughout the reverse diffusion process. Second, to resolve the projection ambiguity and lack of volumetric consistency inherent in 2D alignment, we design the \textbf{Inherent Motional Guider (IMG)}. Derived from shape-agnostic SMPL articulation, IMG encodes motion parameters into an identity-agnostic semantic vector. Injected into the intermediate layers of the DiT, this design provides view-consistent and anatomically-grounded guidance, ensuring that the motion remains physically plausible regardless of character-specific geometry.

Furthermore, to address the lack of standardized protocols for misaligned scenarios, we establish \textbf{CIM-Bench}, a benchmark for cross-identity animation. CIM-Bench is constructed via a high-fidelity motion retargeting pipeline and rigorous manual curation, featuring diverse video pairs with natural structural discrepancies. Extensive experiments demonstrate that RASA significantly outperforms state-of-the-art methods, exhibiting superior robustness in cross-identity settings. In summary, our contributions are: 
(1) \textbf{RASA}, a hierarchical framework for cross-identity animation that systematically disentangles spatial-motional priors; 
(2) A dual-injection mechanism featuring \textbf{SPC} for continuous 2D spatial calibration and \textbf{IMG} for intermediate, identity-agnostic 3D semantic guidance;
(3) \textbf{CIM-Bench}, a standardized benchmark with meticulous curation for evaluating motion fidelity under realistic structural misalignment.
\section{Related Work}

\subsection{Diffusion for Video Generation}
Diffusion models~\cite{ho2020denoising, song2020denoising} alleviate training instability and mode collapse issues common in generative adversarial networks (GANs)~\cite{goodfellow2014generative}, and have become the dominant paradigm for video generation due to their stable optimization and strong visual fidelity~\cite{guo2023animatediff, zhou2022magicvideo, shi2025motionstone}. 

Building upon large-scale text-to-image (T2I) diffusion models~\cite{rombach2022high, singer2022make, chai2023stablevideo, yang2024cogvideox, xu2024easyanimate, cai2025hidreami1highefficientimagegenerative, cai2026hidreamo1imagenativelyunifiedimage}, recent works extend diffusion frameworks to the temporal domain. Early approaches extend pretrained 2D architectures with temporal modules (e.g., temporal convolutions or attention layers) to model inter-frame coherence while preserving spatial priors~\cite{blattmann2023align}. More recently, Diffusion Transformers (DiT)~\cite{peebles2023scalable} replace conventional U-Net backbones with scalable transformer architectures, improving flexibility for long-sequence and high-resolution video synthesis.

To enhance controllability, structured conditions such as depth maps, sketches, human poses, and camera viewpoints are commonly integrated into diffusion pipelines~\cite{li2025tokenmotion, he2024cameractrl, fu20243dtrajmaster, li2024drivingdiffusion}. For instance, ControlNet~\cite{zhang2023adding} introduces condition-specific branches to enforce structural constraints, while ControlNeXt~\cite{peng2024controlnext} adopts a lightweight design with cross-normalization for efficient conditional modulation. In this work, we leverage the strong motion modeling capability of Wan2.1~\cite{wan2025wan}. Unlike existing methods that treat pose solely as a localized condition, we integrate pose signals through a dual-injection mechanism to enhance both spatial grounding and 3D motion perception.

\subsection{Pose-Driven Personalized Generation}
Pose-driven personalized generation aims to transfer motion from a driving sequence to a reference image while preserving character appearance. Recent diffusion-based approaches have significantly improved visual realism and motion fidelity for this task~\cite{tu2025stableanimator,xu2024magicanimate,luo2025dreamactor,cheng2025wan,hu2024animate,hu2025animate}.

Despite these advances, most existing methods assume structural compatibility between the driving pose and the reference image. Directly binding raw pose signals with appearance features often causes geometric distortions when discrepancies in scale or body proportion arise, particularly in cross-identity scenarios. To alleviate structural misalignment, UniAnimate~\cite{wang2025unianimate} performs coarse normalization via estimated offsets, while Champ~\cite{zhu2024champ} leverages SMPL-based parametric alignment. Other strategies include pose modulation~\cite{zhang2025steadydancer}, data augmentation~\cite{tan2024animate}, and learnable similarity transformations~\cite{tu2025stableanimator++}.

In contrast to prior approaches that rely on heuristic preprocessing or late-stage feature fusion, our proposed {RASA} addresses structural misalignment from two complementary perspectives. We introduce the {Spatial Prior Calibrator (SPC)} to neutralize 2D geometric discrepancies at the generative onset, and the {Inherent Motional Guider (IMG)} to provide anatomically-consistent 3D semantic guidance within the intermediate transformer layers.

\section{Method}

\subsection{Preliminaries}

\textbf{Flow Matching for Video Generation.} Recent video generation frameworks commonly adopt Flow Matching (FM) to model the transformation between data and noise distributions. Given a video sample $x \in \mathbb{R}^{T \times H \times W \times 3}$, a pretrained VAE encodes it into the latent representation $z_0=\mathcal{E}(x)$, while $z_1\sim\mathcal{N}(0,I)$ denotes Gaussian noise. Following the optimal transport formulation, FM defines a linear interpolation path: 

\begin{equation}
z_t = (1 - t)z_0 + t z_1,
\end{equation}
where $t\in[0,1]$. The corresponding velocity field is $dz_t/dt = z_1 - z_0.$

\begin{figure}[t]
  \centering
  \includegraphics[width=\linewidth]{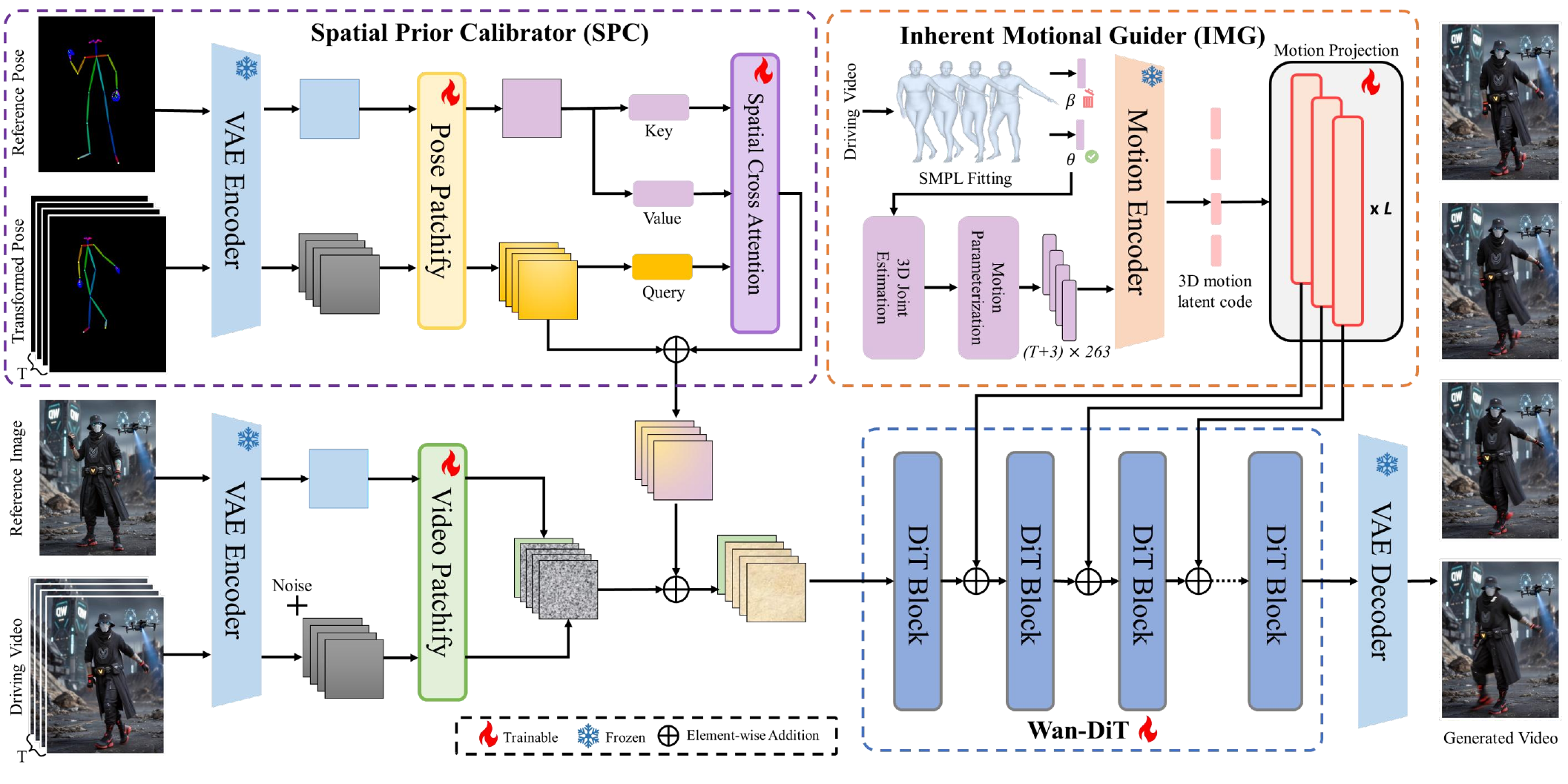}
  \caption{The overview of RASA. Given a reference image and its extracted pose, a driving video and its corresponding pose sequence, as well as SMPL-based motion parameters, we first simulate structural misalignment by applying pose transformations to the driving poses. The motion parameters of the first frame are replicated three times to align temporally with the video latents. Through the Spatial Prior Calibrator module and the Inherent Motional Guider module, motion information is accurately transferred to the reference image while preserving structural consistency.}
\label{network}
\end{figure}

A network $v_\theta(z_t,t,c)$ conditioned on timestep $t$ and control signals $c$ is trained to predict this velocity field using:
\begin{equation}
\mathcal{L} = \mathbb{E}_{z_0, z_1, t}\left[\|v_\theta(z_t, t, c) - (z_1 - z_0)\|_2^2\right].
\end{equation}

During inference, generation starts from Gaussian noise $z_1$ and recovers the clean latent by solving the following Ordinary Differential Equation (ODE):
\begin{equation}
dz_t = v_\theta(z_t, t, c)dt.
\end{equation}

\textbf{Diffusion Transformer Backbone.} To model spatio-temporal dynamics, $v_\theta$ is instantiated as a Diffusion Transformer (DiT). The latent tensor is partitioned into non-overlapping spatio-temporal patches and projected into token embeddings, which are processed by a sequence of Transformer blocks with multi-head self-attention. To capture spatial-temporal relationships, 2D Rotary Positional Encoding (RoPE) is incorporated into the attention computation. In our framework, the DiT backbone takes the geometrically calibrated latent generated by SPC and incorporates 3D semantic guidance through intermediate layers to predict the velocity field for coherent character animation.

\subsection{Overall Framework}

As illustrated in Fig.~\ref{network}, {RASA} is designed to synthesize temporally coherent and photorealistic character videos by leveraging a hierarchical conditioning mechanism. The framework is built upon a Diffusion Transformer (DiT) backbone, which we augment with two synergistic components: the {Spatial Prior Calibrator (SPC)} module for 2D geometric calibration and the {Inherent Motional Guider (IMG)} module for anatomically-grounded motion modeling. Together, these modules enable robust animation even when the reference character and driving motion exhibit severe structural discrepancies.

\paragraph{Reference Identity Conditioning.} 
Given a reference image $I_r$, we first project it into a latent representation $z_r$ via a pretrained VAE encoder. Following the "reference-as-frame" paradigm~\cite{zhou2025realisdance}, $z_r$ is concatenated with the noisy video latents $z_t$ along the temporal dimension. To effectively disentangle static identity from temporal evolution, we apply a significant positional encoding offset to $z_r$. This ensures that while the self-attention layers within the DiT blocks can globally attend to the reference's appearance and texture, the model maintains a clear distinction between the invariant identity and the dynamic latent flow, facilitating superior identity preservation across the generated sequence.

\paragraph{Hierarchical Geometry and Motion Injection.} 
The core of RASA lies in its dual-stream conditioning strategy, which addresses misalignment at different levels of the generative process:

\begin{enumerate}
    \item \textbf{Geometric Rectification at the Generative Onset:} To handle 2D spatial mismatches, we simulate misalignment during training via stochastic spatial perturbations $\mathcal{T}(P_d)$. The SPC module performs structural calibration between the driving pose $\tilde{P}_d$ and the reference pose $P_r$, producing an aligned representation $z_d^{\text{align}}$. This signal is element-wise added to the noisy latent at each denoising step. By continuously enforcing a geometrically consistent structural prior throughout the reverse diffusion process, SPC effectively mitigates scale and skeletal discrepancies before they accumulate into visible appearance distortions.
    
    \item \textbf{Semantic Guidance in Intermediate Layers:} To resolve projection ambiguity and enhance 3D perception, we extract the SMPL-based motion vector $\mathbf{m}$ and project it into the IMG latent $z_m^{3D}$. Unlike the spatial calibration of SPC, $z_m^{3D}$ provides high-level motion semantics. This prior is {hierarchically injected into the intermediate DiT blocks} through learnable projection layers. This multi-depth conditioning reinforces the model's 3D awareness, ensuring that the motion remains anatomically plausible and viewpoint-consistent throughout the entire latent evolution.
\end{enumerate}

By synergizing continuous 2D calibration with deep-stage 3D guidance, RASA achieves a robust decoupling of motion, geometry, and identity, setting a new standard for cross-identity character animation.

\subsection{Spatial Prior Calibrator (SPC)}

While existing approaches attempt to mitigate pose--reference misalignment by heuristic preprocessing, data augmentation, or late-stage feature fusion~\cite{wang2025unianimate,zhu2024champ,tan2024animate,tu2025stableanimator++,zhang2025steadydancer}, they often fail to resolve the fundamental structural discrepancies during the generative process. In contrast, we introduce the \textbf{Spatial Prior Calibrator (SPC)} module, which formulates alignment as a learnable structural calibration process in the latent space prior to DiT conditioning. By shifting the alignment task to the generative onset, SPC provides a structurally consistent prior that anchors the subsequent denoising process, effectively decoupling geometric rectification from appearance synthesis.

\paragraph{Synthetic Misalignment Strategy.} 
To empower the model with robust alignment capabilities, we simulate realistic cross-identity discrepancies during training. Given a driving pose sequence $P_d$, we apply a set of random spatial perturbations $\mathcal{T}$—including stochastic scaling, translation, and proportional skeletal dropout—to obtain a distorted pose sequence $\tilde{P}_d = \mathcal{T}(P_d)$. Our objective is to learn a mapping $\mathcal{R}$ that projects these perturbed poses into a reference-consistent structural space conditioned on the reference pose $P_r$:
\begin{equation}
P_d^{\text{align}} = \mathcal{R}(\tilde{P}_d, P_r).
\end{equation}

\paragraph{Feature-Level Correspondence Modeling.} 
Specifically, we utilize a shared light-weight encoder to project $P_r$ and $\tilde{P}_d$ into a latent space, yielding a static reference latent $z_r \in \mathbb{R}^{C \times H \times W}$ and a temporal driving latent $z_d \in \mathbb{R}^{T \times C \times H \times W}$. To explicitly model the non-local spatial correspondence between the two potentially disparate skeletons, we employ a {frame-wise spatial cross-attention} mechanism. 

For each frame $t$, the driving latent $z_{d,t}$ is flattened into a sequence of tokens $Q_t \in \mathbb{R}^{N \times C}$ (where $N = H \times W$), while the reference latent $z_r$ serves as the shared key $K$ and value $V$. To preserve the 2D structural topology essential for skeletal integrity, we incorporate \textbf{2D Rotary Positional Encoding (RoPE)} into the attention computation. This enables the model to adaptively warp and calibrate the driving features based on the reference's geometric constraints. The aligned feature for frame $t$ is formulated as:
\begin{equation}
\hat{z}_{d,t} = z_{d,t} + \text{Attention}\left(\text{RoPE}(Q_t), \text{RoPE}(K), V\right).
\end{equation}

\paragraph{Structural Condition Injection.} 
The resulting aligned features $\hat{z}_d$ are reshaped back to the spatial-temporal dimensions $\mathbb{R}^{T \times C \times H \times W}$. Then, this representation is element-wise added to the noisy latent representation at the beginning of each denoising step of the DiT backbone. By introducing structurally calibrated cues before each denoising iteration, SPC continuously provides a geometrically rectified structural prior throughout the reverse diffusion process. This persistent guidance ensures that the target identity remains anchored to the correct skeletal proportions and global position during latent evolution, effectively mitigating structural drift and accumulated misalignment errors.

\subsection{Inherent Motional Guider (IMG)}
While SPC effectively harmonizes geometric discrepancies in the 2D spatial domain, 2D pose representations are inherently view-dependent and suffer from projection ambiguity. Such limitations often lead to a lack of volumetric consistency, particularly during complex articulations where depth information is crucial for maintaining skeletal integrity. To complement the 2D alignment, we introduce the \textbf{Inherent Motional Guider (IMG)} module. By leveraging a compact, SMPL-based motion representation, IMG encodes global body kinematics into a latent space, providing the DiT backbone with anatomically-grounded guidance that is invariant to character-specific geometry.

\paragraph{Shape-Agnostic Motion Representation.} 
Unlike previous works~\cite{zhu2024champ, ding2025mtvcrafter} that utilize explicit 3D joint coordinates—which often entangle motion with the subject's specific skeletal scale—our IMG aims for a purely decoupled motion descriptor.
We utilize ScoreHMR~\cite{stathopoulos2024score} to estimate the 3D human mesh parameters from the driving video. To ensure strict geometry invariance, we explicitly discard the subject-specific shape parameters ($\beta$) and construct the motion prior solely from the articulation parameters ($\theta$). Following the standard parameterization in motion synthesis~\cite{guo2022generating}, we represent the articulated sequence as a spatially-aware motion vector $\mathbf{m} \in \mathbb{R}^{T \times 263}$, which encapsulates root-relative joint rotations, velocities, and foot contact signals. This representation ensures that motion signal remains identity-agnostic and focused exclusively on the intrinsic dynamics.

\paragraph{Hierarchical Motion Injection.} 
To transform the raw motion vector into a semantically rich condition, we employ a pretrained motion encoder $\mathcal{E}_m$~\cite{zhang2023t2m} to project $\mathbf{m}$ into a latent sequence $z_m^{3D} \in \mathbb{R}^{T \times D}$. Recognizing that 3D motion dynamics represent a high-level semantic signal—analogous to phonetic features in audio-driven generation~\cite{gan2025omniavatar}—we design a hierarchical injection strategy. 

Specifically, we introduce a set of $L=14$ learnable motion projection layers. Rather than affecting the initial spatial layout, the motion latent $z_m^{3D}$ is injected into the \textbf{intermediate blocks} (specifically, blocks 2 through 15) of the DiT backbone. 
This design choice is deliberate: while SPC (at the generative onset) rectifies the coarse 2D geometry , the IMG (in the intermediate layers) provides continuous, depth-aware refinement as the latent features evolve. This synergy ensures that the generated character not only adheres to the reference's 2D proportions but also obeys 3D physical and anatomical constraints throughout the denoising process.

\subsection{Cross-Identity Misalignment Benchmark}

Our method explicitly addresses structural misalignment between reference images and driving pose sequences. However, existing benchmarks provide limited support for systematically evaluating authentic cross-identity discrepancies. For instance, MisAlign100~\cite{tu2025stableanimator++} simulates misalignment by applying random linear transformations (e.g., rotation, scaling, and translation) to driving poses extracted from the \textit{same subject and video} as the reference image. While this strategy introduces synthetic geometric perturbations, it does not fully reflect the physical realism of cross-identity animation.

\begin{wrapfigure}{r}{0.5\textwidth}
\vspace{-7.5mm}
  \begin{center}
\includegraphics[width=1.0\linewidth,clip=True]{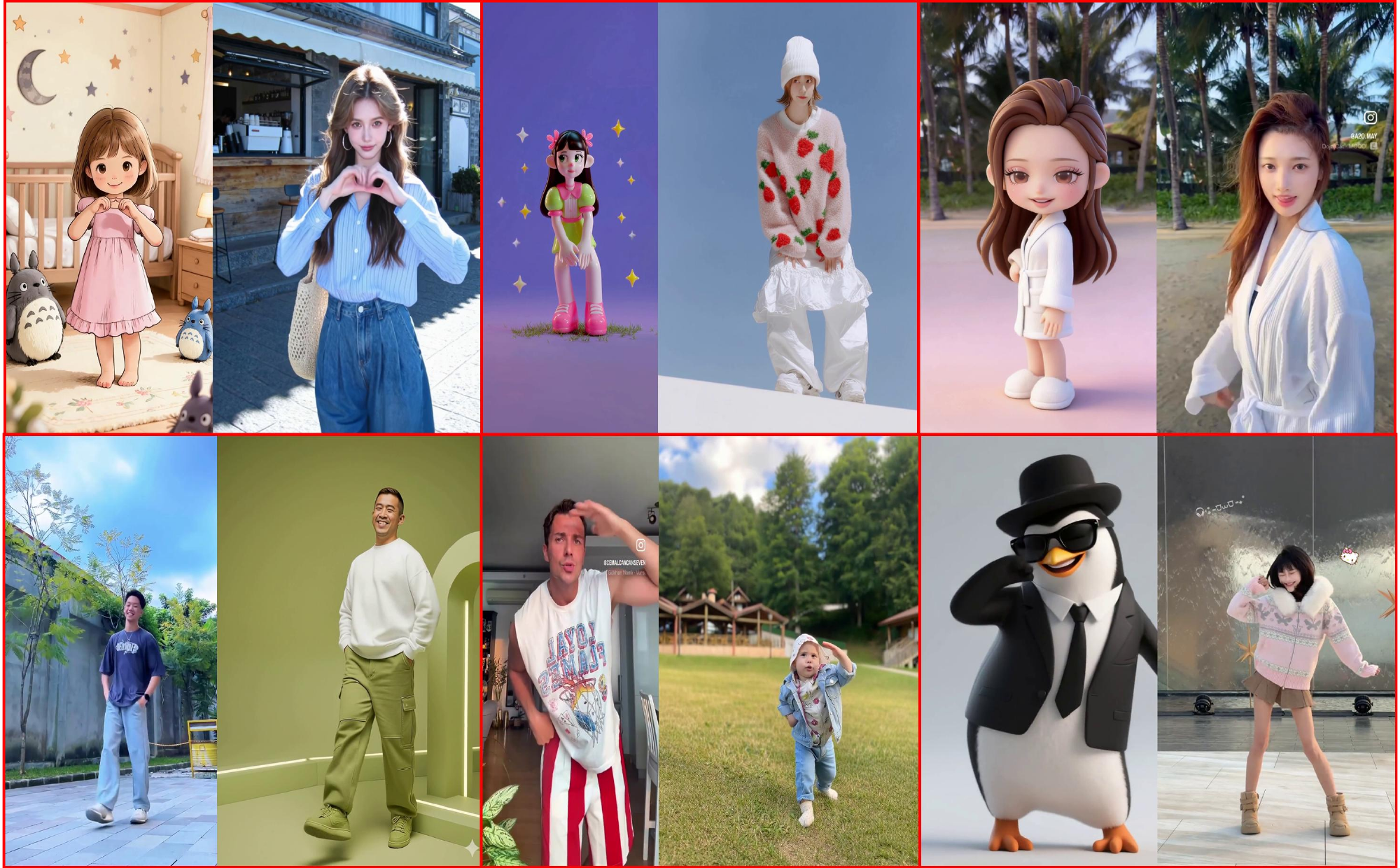}
  \end{center}
\caption{Illustration of CIM-Bench. Different identities perform the same motion sequence, creating authentic cross-identity structural discrepancies.}\label{cim_demo}
\vspace{-5.5mm}
\end{wrapfigure}

To bridge this evaluation gap and provide a rigorously realistic testbed, we construct a novel benchmark termed \textbf{CIM-Bench}, which incorporates both real human subjects and stylized synthetic characters performing identical motion patterns (see Fig.~\ref{cim_demo}). Specifically, we use 
a state-of-the-art text-to-image diffusion model 
to generate diverse stylized character identities and collect real human portraits from online sources to form paired images. For each identity pair, we employ 
a commercial video synthesis backbone 
to synthesize videos in which different characters execute the same motion sequence.

Since pose estimation from real human videos is generally more stable and reliable, we extract pose sequences from these videos as driving motions. The extracted poses are then applied to the corresponding stylized character videos, thereby constructing structurally misaligned cross-identity scenarios. This design preserves natural motion dynamics while introducing realistic structural discrepancies in body shape, proportion, and appearance across identities, enabling a more faithful evaluation of a model's ability to handle structural misalignment while maintaining robust and stable motion transfer.

\section{Experiments}
\subsection{Experimental Settings}
\paragraph{Implementation Details.}
Our framework is built upon the open-source text-to-video diffusion backbone Wan2.1 (1.3B)~\cite{cheng2025wan}. 
For training, we leverage several publicly available datasets, including TikTok~\cite{jafarian2021learning}, Champ~\cite{zhu2024champ}, and UBC~\cite{zablotskaia2019dwnet}. In addition, we collect approximately 5,000 human-centric videos from the internet to further enrich the training data. All videos are resized to a spatial resolution of $832 \times 480$, and pose sequences are uniformly extracted using DWPose~\cite{yang2023effective}. 
During training, we randomly sample clips of 41 consecutive frames to encourage temporal coherence learning, with the first frame of each clip selected as the reference image. 
Following Animate-X~\cite{tan2024animate}, we apply pose perturbation transformations to the driving poses to simulate structural misalignment. The model is trained on 8 NVIDIA A100 GPUs using the AdamW optimizer with a learning rate of $1\times10^{-5}$, and the full training process takes approximately two days. For the construction of CIM-Bench, we enforce strict pose quality control through manual filtering, removing samples where DWPose fails to produce clear or temporally consistent keypoints. The final benchmark contains 656 high-quality character videos paired with corresponding pose sequences.

\paragraph{Evaluation Data and Metrics.}
We evaluate our method on the TikTok dataset~\cite{jafarian2021learning} following the experimental protocol of prior work~\cite{wang2024disco,ding2025mtvcrafter,chang2023magicpose}. 
We further conduct cross-identity evaluation on CIM-Bench by sampling a subset for testing, while the remaining samples are excluded from training to ensure a fair evaluation. For quantitative evaluation, we adopt both image-level and video-level metrics. The image-level metrics include PSNR~\cite{hore2010image}, SSIM~\cite{wang2004image}, LPIPS~\cite{zhang2018unreasonable}, and FID~\cite{heusel2017gans}. The video-level metrics include FVD~\cite{unterthiner2018towards} and FID-VID~\cite{balaji2019conditional}, which assess overall visual fidelity and temporal coherence. In addition, we employ Sim-Arc~\cite{deng2019arcface} and Face-FID to evaluate facial identity consistency. To assess computational efficiency, we additionally report inference latency and GPU memory consumption during inference.

\begin{table}[t]
\centering
\caption{Quantitative results on the TikTok~\cite{jafarian2021learning} dataset.}
\label{tab:tiktok_test}
\begin{tabular}{lcccccc}
\toprule
\multirow{2}{*}{Method} 
& \multicolumn{4}{c}{Image Metrics} 
& \multicolumn{2}{c}{Video Metrics}  \\

\cmidrule(r){2-5} \cmidrule(r){6-7}

& PSNR~$\uparrow$ 
& SSIM~$\uparrow$ 
& LPIPS~$\downarrow$ 
& FID~$\downarrow$
& FVD~$\downarrow$ 
& FID-VID~$\downarrow$ \\
\midrule
DreamPose~\cite{karras2023dreampose} & 12.82 & 0.511 & 0.442 & 72.62 & 551.02 & 78.77 \\
MagicAnimate~\cite{xu2024magicanimate} & - & 0.714 & 0.239 & 32.09 & 179.07 & 21.75 \\
Champ~\cite{zhu2024champ} & - & 0.802 & 0.234 & - & \underline{160.82} & 21.07 \\

DisCo~\cite{wang2024disco} & 16.55 & 0.668 & 0.292 & 30.75 & 292.80 & 59.90 \\
Animate Anyone~\cite{hu2024animate} & 17.40 & 0.734 & 0.287 & 60.70 & 453.00 & 60.00 \\
MimicMotion~\cite{zhang2024mimicmotion} & 20.10 & 0.795 & 0.232 & 35.62 & 594.00 & 9.30 \\
ControlNeXt~\cite{peng2024controlnext} & 18.52 & 0.763 & 0.246 & 39.66 & 398.32 & 24.29 \\
StableAnimator~\cite{tu2025stableanimator} & 18.12 & 0.771 & 0.257 & 40.17 & 253.69 & 22.79 \\
SteadyDancer~\cite{zhang2025steadydancer} & 17.67 & 0.749 & 0.263 & 30.65 & 451.30 & - \\
Human-DiT~\cite{gan2025humandit} & \underline{20.50} & \underline{0.815} & 0.220 & 41.60 & 237.00 & 24.30 \\
One-to-All-1.3B~\cite{shi2025one} & 17.75 & 0.788 & 0.269 & 74.96 & 361.85 & 21.37 \\
One-to-All-14B~\cite{shi2025one} & 18.07 & 0.812 & 0.254 & 50.49 & 297.94 & 13.93 \\
MTVCrafter~\cite{ding2025mtvcrafter} & 19.37 & 0.784 & \underline{0.217} & \underline{19.46} & \textbf{140.60} & \underline{6.98} \\

\midrule
\textbf{Ours} & \textbf{22.03} & \textbf{0.830} & \textbf{0.189} & \textbf{18.93} & 190.82 & \textbf{3.32} \\

\bottomrule
\end{tabular}
\end{table}

\begin{table}[t]
\centering
\caption{Quantitative results on the CIM-Bench.}
\label{tab:CIM_test}
\resizebox{\linewidth}{!}{%
\begin{tabular}{lcccccccccc}
\toprule
\multirow{2}{*}{Method} 
& \multicolumn{4}{c}{Image Metrics} 
& \multicolumn{2}{c}{Video Metrics}  
& \multicolumn{2}{c}{Identity Metrics}
& \multicolumn{2}{c}{Efficiency}\\
\cmidrule(r){2-5} \cmidrule(r){6-7} \cmidrule(r){8-9} \cmidrule(r){10-11}
& PSNR~$\uparrow$ 
& SSIM~$\uparrow$ 
& LPIPS~$\downarrow$ 
& FID~$\downarrow$
& FVD~$\downarrow$ 
& FID-VID~$\downarrow$ 
& Sim-Arc~$\uparrow$
& Face-FID~$\downarrow$ 
& Latency (s)
& VRAM (MiB)
\\
\midrule
Animate-X~\cite{tan2024animate} & 14.05 & 0.426 & 0.474 & 105.14 & \underline{1312.60} & \underline{75.47} & 0.51 & \underline{146.55} & 109 & 14,263 \\
MTVCrafter~\cite{ding2025mtvcrafter} & \underline{14.19} & \underline{0.431} & \underline{0.457} & \underline{85.21} & 1723.89 & 85.21 & \underline{0.54} & 154.77 & 170 & 20,680\\
Wan-Animate~\cite{cheng2025wan} & 11.29 & 0.372 & 0.585 & 178.58 & 1920.88 & 137.05 & 0.35 & 225.29 & 327 & 42,291\\
One-to-All-1.3B~\cite{shi2025one} & 13.75 & 0.383 & 0.471 & 96.70 & 1324.74 & 76.79 & 0.51 & 163.99 & 91 & 21,880\\
\midrule
\textbf{Ours} & \textbf{15.93} & \textbf{0.472} & \textbf{0.372} & \textbf{60.96} & \textbf{667.96} & \textbf{17.63} & \textbf{0.72} & \textbf{90.63} & 61 & 11,951\\
\bottomrule
\end{tabular}%
}
\end{table}

\subsection{Comparison with State-of-the-Art Methods}

\paragraph{Quantitative Results.}
We compare our method with recent state-of-the-art human animation approaches on both the TikTok dataset and CIM-Bench, as shown in Tab.~\ref{tab:tiktok_test} and Tab.~\ref{tab:CIM_test}. Under the self-driven evaluation protocol on TikTok, our method achieves the best performance across all image-level metrics, demonstrating superior visual fidelity and identity preservation. For video-level evaluation, although our FVD is slightly higher than that of MTVCrafter~\cite{ding2025mtvcrafter} and Champ~\cite{zhu2024champ}, MTVCrafter is built upon the large-scale CogVideoX-5B~\cite{yang2024cogvideox} backbone, which benefits from substantially greater generative capacity due to its parameter scale. 
\begin{figure}[hbtp]
  \centering
  \includegraphics[width=\linewidth]{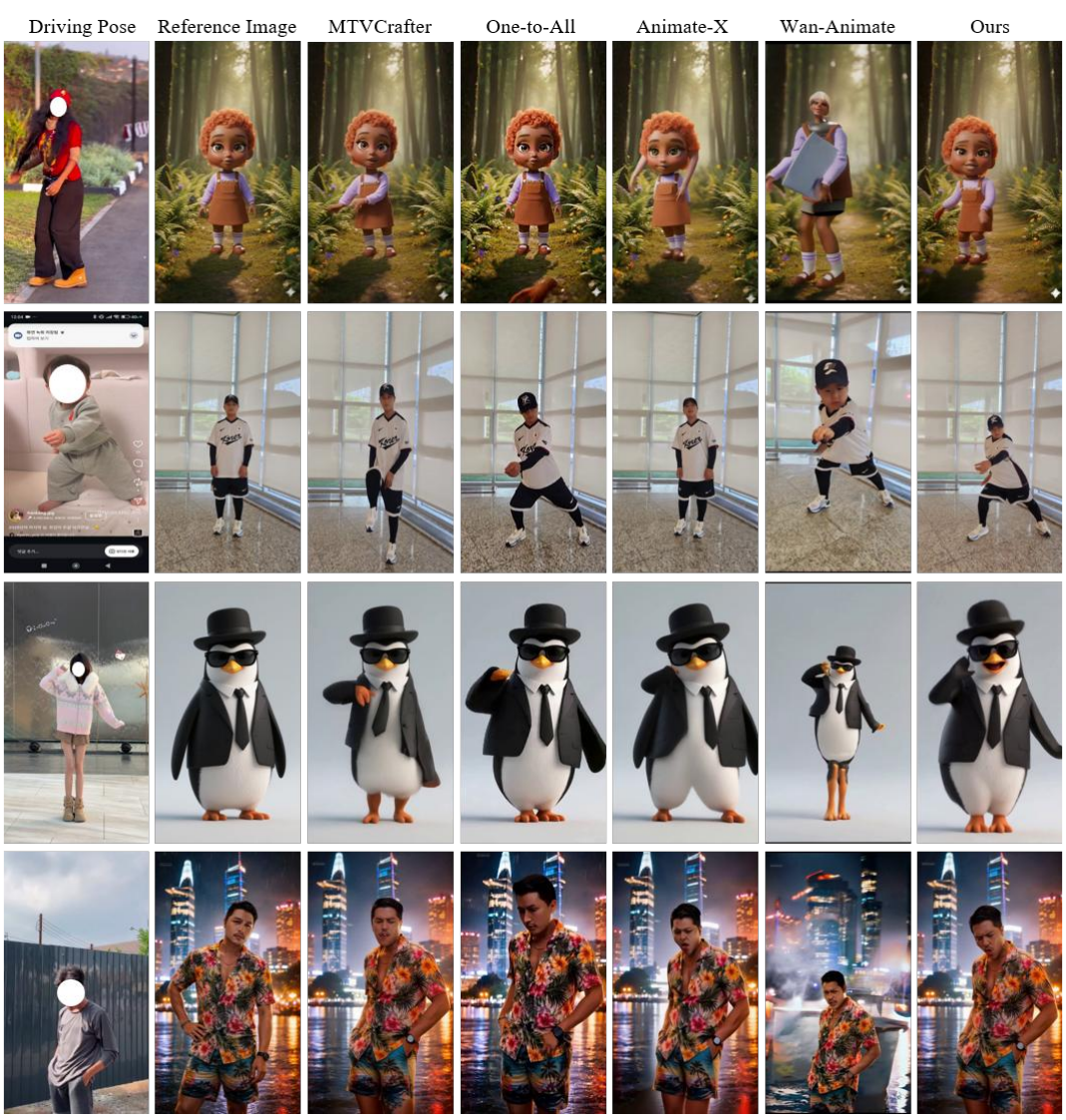}
  \caption{Qualitative comparison for cross-identity character animation task under structurally misaligned scenarios.}
\label{img:qualitative}
\end{figure}
Meanwhile, Champ requires five distinct motion representations as conditioning signals for generation, leading to higher model complexity. In contrast, our method adopts a lightweight 1.3B backbone and requires only 2D poses and motion parameters as conditioning signals. Despite this gap in model size and conditioning complexity, our method remains highly competitive and surpasses larger-scale models such as One-to-All-14B~\cite{shi2025one} and Human-DiT~\cite{gan2025humandit} on multiple metrics. On the more challenging CIM-Bench, which introduces substantial pose–reference misalignment, our method consistently outperforms all competing approaches across both image-level and video-level metrics. Moreover, the superior identity metrics demonstrate its ability to maintain strong facial identity consistency with the reference character. These results demonstrate the effectiveness of our structural alignment and motion modeling design in handling pose--reference misalignment, enabling temporally coherent and stable animation generation with competitive inference latency and low GPU memory consumption.

\paragraph{Qualitative comparisons.}
Fig.~\ref{img:qualitative} presents qualitative results under cross-identity and pose–reference misalignment scenarios. MTVCrafter~\cite{ding2025mtvcrafter} and Animate-X~\cite{tan2024animate} exhibit noticeable deficiencies in motion fidelity, including inaccurate motion transfer, positional shifts, and structural inconsistencies in the generated body. Although One-to-All~\cite{shi2025one} and Wan-Animate~\cite{hu2025animate} generate relatively accurate motion trajectories, they suffer from identity distortion when substantial structural discrepancies exist between the reference and driving poses. Moreover, due to the absence of an explicit structural alignment mechanism, Wan-Animate introduces spatial displacement artifacts, leading to positional shifts of the generated character. In contrast, our method accurately follows the driving pose while preserving strong identity consistency with the reference image. The generated animations exhibit stable spatial structure and consistent appearance retention, demonstrating the effectiveness of our spatial prior calibrator and inherent motional guide modules.

\subsection{Ablation Study}
\begin{figure}[h]
  \centering
  \includegraphics[width=\linewidth]{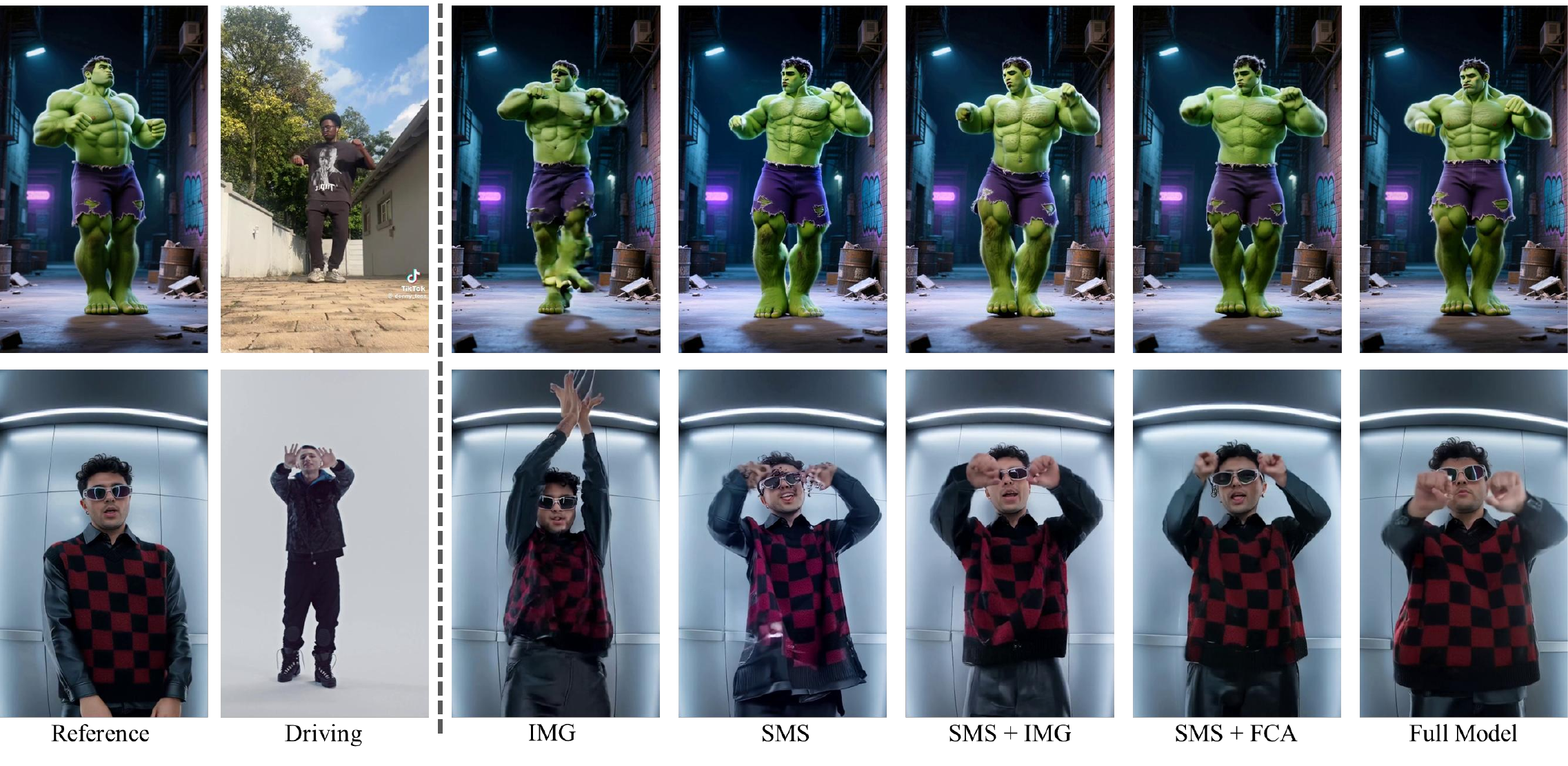}
  \caption{Qualitative results of ablation study.}
\label{img:ablation_test}

\end{figure}

We conduct comprehensive ablation studies on the TikTok and CIM-Bench datasets to evaluate the individual contributions of our core components: the synthetic misalignment strategy (SMS), the frame-wise cross-attention (FCA) within the SPC module, and the inherent motional guider (IMG). Quantitative results are summarized in Tab.~\ref{tab:ablation}, and corresponding qualitative comparisons are presented in Fig.~\ref{img:ablation_test}.

\begin{table}[hbtp]
\centering
\small
\caption{Ablation study on each key component of our model on TikTok~\cite{jafarian2021learning}/CIM-Bench. SMS denotes performing random spatially transformation on driving poses. FCA denotes performing frame-wise cross attention on reference and driving pose. IMG denotes performing inherent motional guider.}

\label{tab:ablation}

\resizebox{\columnwidth}{!}{
\begin{tabular}{ccccccccc}
\toprule
SMS & FCA & IMG 
& PSNR $\uparrow$ 
& SSIM $\uparrow$ 
& LPIPS $\downarrow$ 
& FID $\downarrow$
& FVD $\downarrow$ 
& FID-VID $\downarrow$ \\
\midrule

\cmark &  & & 18.90/14.60 & 0.799/0.453 & 0.209/0.393 & 23.04/80.96 & 316.45/737.48 & 10.77/30.93 \\

 &  & \cmark & 15.47/13.89 & 0.707/0.416 & 0.356/0.445 & 41.57/91.17 & 1567.03/1625.69 & 15.90/37.46 \\

\cmark &  & \cmark & 18.51/14.14 & 0.811/0.470 & 0.238/0.399 & 28.83/77.34 & 487.66/706.22 & 22.06/34.51 \\

\cmark & \cmark &  & 20.86/15.26 & 0.820/0.471 & 0.202/0.391 & 22.01/71.16 & 289.16/686.79 & 6.56/27.05 \\

\cmark & \cmark & \cmark & \textbf{22.03/15.93} & \textbf{0.830/0.472} & \textbf{0.189/0.372} & \textbf{18.93/60.96} & \textbf{190.82/667.96} & \textbf{3.32/17.63} \\

\bottomrule
\end{tabular}}
\end{table}

\textbf{Effectiveness of SPC.} 
As reported in Tab.~\ref{tab:ablation}, relying solely on spatially transformed driving poses (SMS) yields suboptimal performance, especially under the severe structural discrepancies of CIM-Bench (e.g., yielding a low SSIM of 0.453 and a high FVD of 737.48). Integrating the FCA mechanism (SMS+FCA) significantly boosts generative quality by performing explicit 2D structural calibration. Quantitatively, this yields an improvement in PSNR (from 18.90 to 20.86 on TikTok) and a reduction in FID (from 23.04 to 22.01). Visually, as shown in Fig.~\ref{img:ablation_test}, the model equipped with SPC (SMS+FCA) successfully resolves spatial dislocations and preserves the reference character's skeletal proportions, demonstrating that continuous geometric alignment effectively mitigates appearance distortions and provides reliable structural priors for animation generation.

\textbf{Effectiveness of IMG.} While SPC successfully addresses 2D geometric mismatches, it fundamentally lacks depth awareness. Providing only the IMG without proper spatial conditioning (IMG only) results in the worst image fidelity (PSNR of 15.47 on TikTok), as the implicit cue alone is insufficient to guide fine-grained spatial rendering. However, when integrated with SPC (our Full Model: SMS+FCA+IMG), the IMG module provides anatomically-grounded guidance, leading to the best overall performance. Notably, temporal consistency metrics see substantial gains: FVD drops drastically from 289.16 to 190.82 on TikTok, and FID-VID improves from 6.56 to 3.32. These quantitative leaps confirm that SPC and IMG are highly complementary—SPC ensures accurate 2D spatial alignment, while IMG provides anatomically grounded 3D guidance for temporally coherent and view-consistent motion dynamics.

\textbf{Inference Overhead.} 
Inference is conducted on a single NVIDIA H100 GPU for generating a 41-frame video at $832\times480$ resolution. SPC and IMG introduce only 19.78M parameters (\(<1.52\%\) of the 1.3B backbone) and incur 90.29G and 0.17G FLOPs, respectively. They achieve substantial gains in structural alignment and motion consistency with only a 2-second latency overhead (59s to 61s), demonstrating a favorable efficiency--performance trade-off.

\begin{wrapfigure}{t}{0.5\textwidth}
\vspace{-4.5mm}
  \begin{center}
\includegraphics[width=1.0\linewidth,clip=True]{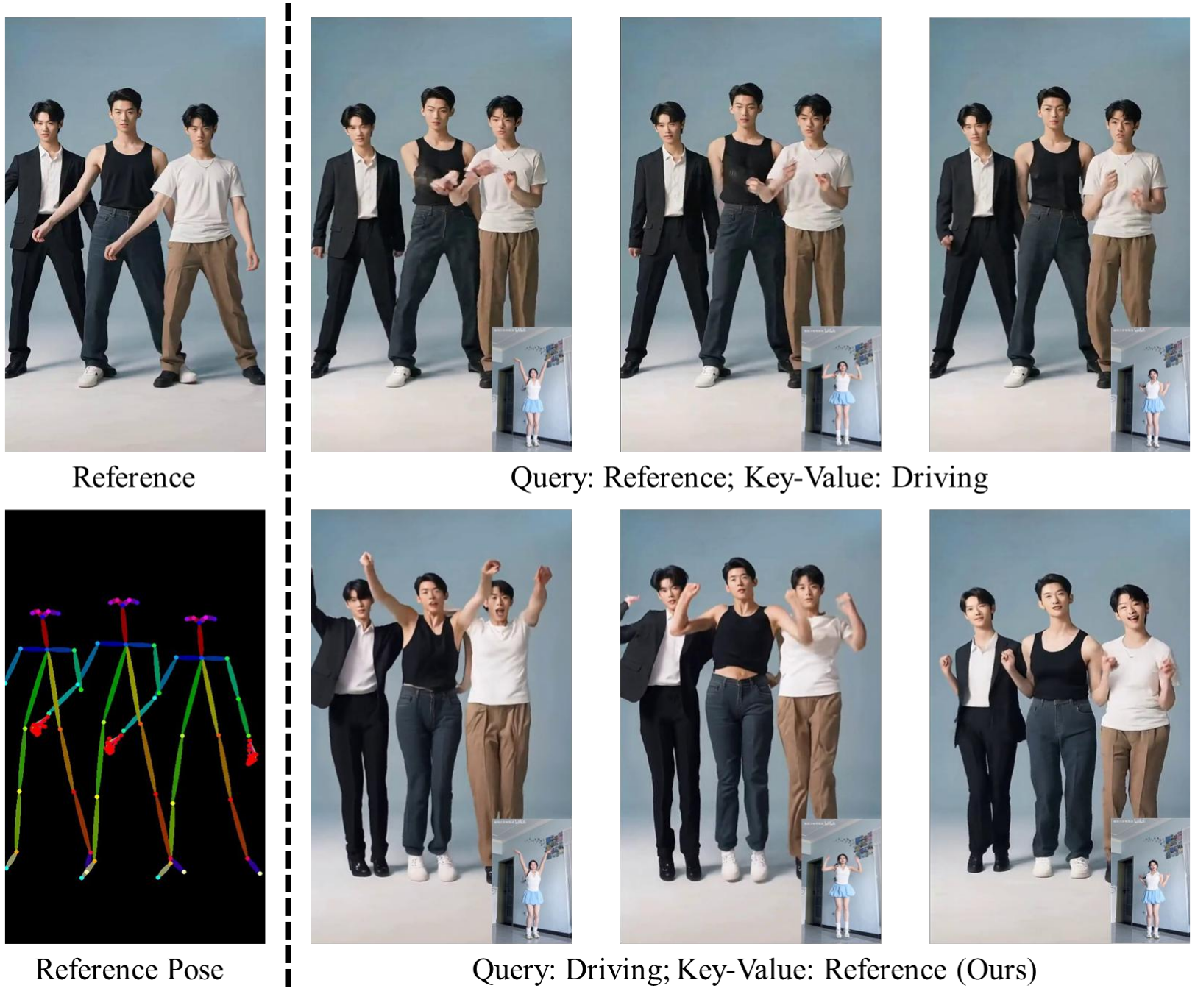}
  \end{center}
\vspace{-4.5mm}
\caption{Ablation study of the FCA attention design in SPC module. Using the driving latent as the query preserves motion transfer, while the reference-query variant weakens motion dynamics.}
\label{img:ablation_map}
\vspace{-5.5mm}
\end{wrapfigure} 

\textbf{Ablation on FCA Attention Formulation.} Furthermore, we explore an alternative attention formulation within the SPC module by swapping the query-key roles: using the reference latent as the query and the driving latent as the key-value pair. As shown in Fig.~\ref{img:ablation_map}, this reverse design fails to produce effective motion transfer. We attribute this to the static nature of the reference image serving as the query, which inadvertently suppresses dynamic motion cues during alignment. Conversely, using the temporally evolving driving latent as the query effectively retrieves and broadcasts the aligned identity features, successfully preserving complex motion patterns.

\begin{figure}[h]
  \centering
  \includegraphics[width=\linewidth]{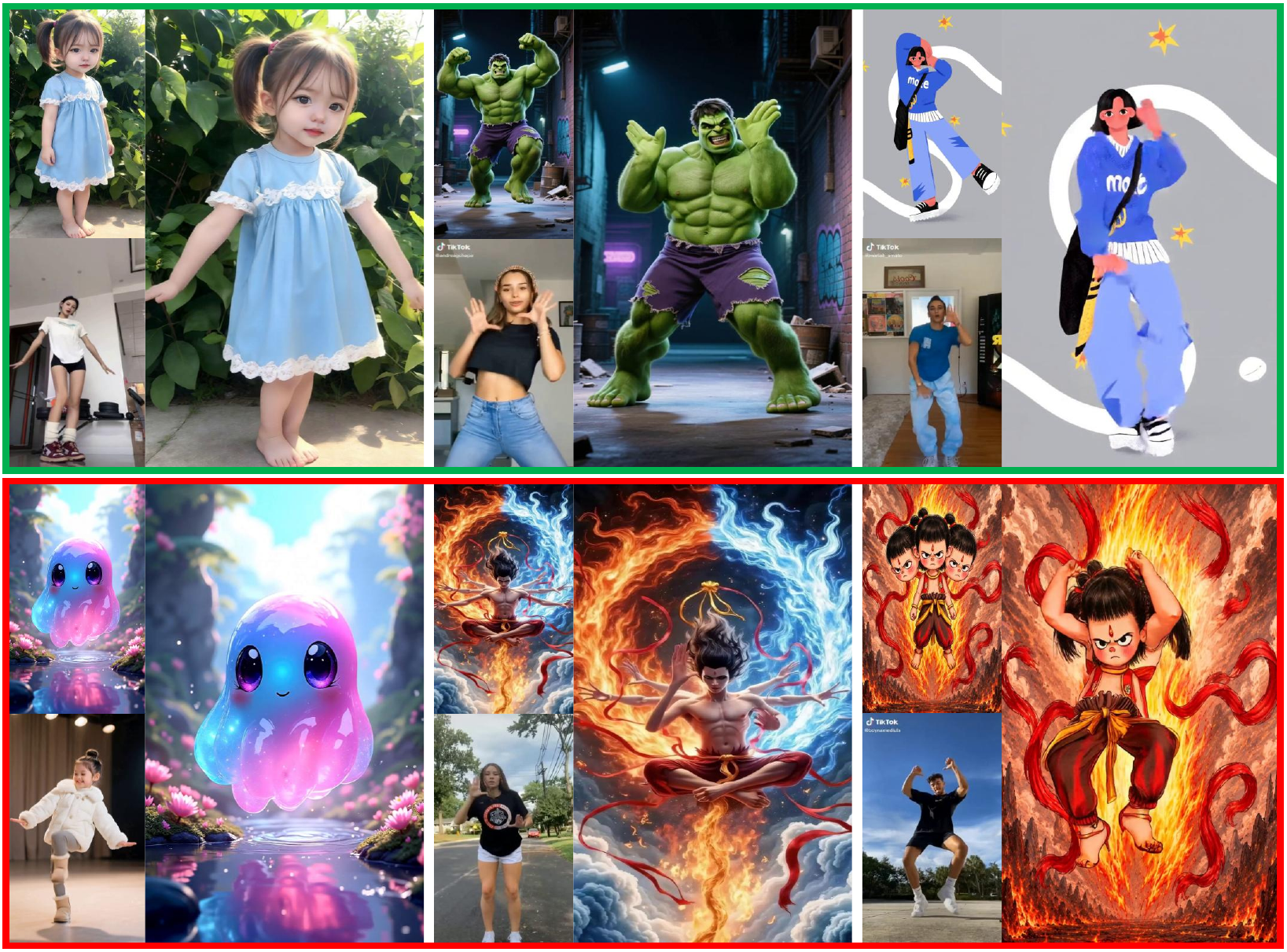}
  \caption{Representative results under extreme proportions and non-humanoid topologies. The green boxes indicate successful cases under extreme body proportions, while the red boxes highlight failure cases involving non-humanoid topologies.}
\label{img:demo_case}
\end{figure}

\section{Conclusion}
In this paper, we present RASA, a unified diffusion framework that addresses the challenging problem of cross-identity character animation. We identify pose-reference misalignment as the core bottleneck in existing methods and tackle it from two complementary perspectives: geometric calibration and semantic motion preservation. For the geometric perspective, our Spatial Prior Calibrator module performs continuous structural alignment throughout the diffusion process, effectively mitigating scale and proportion mismatches. For the semantic perspective, we introduce an Inherent Motional Guider module that extracts shape-agnostic articulation parameters and seamlessly injects them into the diffusion backbone, providing stable, depth-aware motion cues that resolve 2D geometric ambiguity. Furthermore, we establish CIM-Bench, a high-fidelity benchmark explicitly designed to evaluate realistic cross-identity structural variations. 
Extensive experiments on both standard datasets and CIM-Bench demonstrate that RASA achieves SOTA visual fidelity and motion stability, providing a new paradigm for robust cross-identity character animation.

\section{Limitation and Future Work}
Although RASA achieves superior performance over existing methods and successfully handles characters with extreme humanoid proportions, several limitations remain, as shown in Fig.~\ref{img:demo_case}. First, RASA assumes a 1:1 body-part correspondence between the reference character and the driving motion, making it unsuitable for non-humanoid characters with fundamentally different topologies (e.g., multi-headed, multi-armed, or limb-less figures), where reliable spatial correspondences cannot be established. Second, our framework focuses on full-body animation, while hand and facial motions rely solely on pose representations. Since SMPL does not explicitly model facial expressions or hand articulations, the quality of these motions is largely constrained by the estimated 2D poses. Future work will focus on collecting more diverse, high-quality data to further improve robustness and generalization, incorporating explicit facial and hand modeling based on SMPL-X and MANO, and extending the framework beyond the current 1:1 correspondence assumption toward topology-aware animation for a wider range of character categories.

\section*{Acknowledgements}

This work was supported by the Key Science \& Technology Project of Anhui Province (No. 202523o09050002) and the National Natural Science Foundation of China (No. 62372151 and No. 72188101).
%
%
\bibliographystyle{splncs04}
\bibliography{main}
\end{document}